\documentclass[letterpaper, 10 pt, conference]{ieeeconf}  

\IEEEoverridecommandlockouts                              

\usepackage{graphics} 
\usepackage{epsfig} 
\usepackage{mathptmx} 
\usepackage{times} 
\usepackage{caption}
\usepackage{amssymb}
\usepackage{multirow} 
\usepackage[T1]{fontenc}
\usepackage[utf8]{inputenc}
\usepackage{hyperref}
\usepackage{adjustbox} 

\usepackage[T1]{fontenc} 
\usepackage{cite}
\usepackage{booktabs}
\usepackage{makecell}

\usepackage{amsmath,amssymb,amsfonts}
\usepackage{algorithmic}
\usepackage{cleveref} 

\usepackage{setspace}
\usepackage{subfigure}
\usepackage[utf8]{inputenc}
\usepackage{graphicx}
\usepackage{textcomp}
\usepackage[table]{xcolor}
\usepackage{multirow}
\usepackage{threeparttable}
\usepackage{tabularx}
\usepackage{mathtools}
\usepackage[free-standing-units=true]{siunitx}
\usepackage{float}
\usepackage{titlesec}
\titleformat{\subsubsection}[runin]
  {\bfseries}   
  {}            
  {0pt}         
  {}  [.]  
\newcommand{\rom}[1]{\uppercase\expandafter{\romannumeral #1\relax}}

\colorlet{tabfirst}{red!35}
\colorlet{tabsecond}{orange!35}
\colorlet{tabthird}{yellow!35}

\usepackage{array}
\newcommand{\PreserveBackslash}[1]{\let\temp=\\#1\let\\=\temp}
\newcolumntype{C}[1]{>{\PreserveBackslash\centering}p{#1}}
\newcolumntype{R}[1]{>{\PreserveBackslash\raggedleft}p{#1}}
\newcolumntype{L}[1]{>{\PreserveBackslash\raggedright}p{#1}}

\usepackage{xspace}

\definecolor{ourgray}{gray}{0.9}

\makeatletter
\let\NAT@parse\undefined
\makeatother
\author{Lingxiang Hu$^{1}$, 
         Naima Ait Oufroukh$^{1}$, 
         Fabien Bonardi$^{1}$, 
         and Raymond Ghandour$^{2}$
\thanks{$^{1}$Lingxiang Hu, Naima Ait Oufroukh, and Fabien Bonardi are with IBISC, Université Paris-Saclay, 91000 Évry, France. 
        E-mail: \texttt{hulxhlx@gmail.com}, \texttt{naima.aitoufroukh@univ-evry.fr}, \texttt{fabien.bonardi@univ-evry.fr}}
\thanks{$^{2}$Raymond Ghandour is with the College of Engineering and Technology, American University of the Middle East, Kuwait. 
        E-mail: \texttt{Raymond.Ghandour@aum.edu.kw}}
}
\begin{document}
\title{\bf{\LARGE
EC3R-SLAM: Efficient and Consistent Monocular Dense SLAM with Feed-Forward 3D Reconstruction
 \\}
}
\maketitle
\begin{abstract}
The application of monocular dense Simultaneous Localization and Mapping (SLAM) is often hindered by high latency, large GPU memory consumption, and reliance on camera calibration. To relax this constraint, we propose EC3R-SLAM, a novel calibration-free monocular dense SLAM framework  that jointly achieves high localization and mapping accuracy, low latency, and low GPU memory consumption. This enables the framework to achieve efficiency through the coupling of a tracking module, which maintains a sparse map of feature points, and a mapping module based on a feed-forward 3D reconstruction model that simultaneously estimates camera intrinsics. In addition, both local and global loop closures are incorporated to ensure mid-term and long-term data association, enforcing multi-view consistency and thereby enhancing the overall accuracy and robustness of the system. Experiments across multiple benchmarks show that EC3R-SLAM achieves competitive performance compared to state-of-the-art methods, while being faster and more memory-efficient. Moreover, it runs effectively even on resource-constrained platforms such as laptops and Jetson Orin NX, highlighting its potential for real-world robotics applications. We will make our source code publicly available. Project page: \url{https://h0xg.github.io/ec3r/}.

\end{abstract}


\section{Introduction}
\label{sec:intro}

Monocular dense SLAM has found wide applications in autonomous driving, robotic navigation, and AR/VR~\cite{matsuki2024gaussian,maggio2025vggt}. 
Existing approaches exploit neural priors~\cite{murai2024mast3r,maggio2025vggt,teed2021droid,zhang2024glorie,zhang2023go} to recover dense geometry from monocular input. 
Although effective, these methods often suffer from high GPU memory usage~\cite{murai2024mast3r,maggio2025vggt,zhang2024hi}, significant latency~\cite{zhang2024glorie,matsuki2024gaussian,sandstrom2025splat,zhang2024hi}, and complex optimization pipelines~\cite{teed2021droid,matsuki2024gaussian}, which limit their suitability for real-time deployment. 
More recently, a new class of 3D reconstruction models~\cite{wang2024spann3r,wang2024dust3r,duisterhof2024mast3r,wang2025vggt,yang2025fast3r} has demonstrated the ability to recover dense geometry directly from uncalibrated RGB frames. Among them, VGGT~\cite{wang2025vggt} and Fast3R~\cite{yang2025fast3r} are able to reconstruct thousands of images within seconds, 
but their high GPU memory requirements render them unsuitable for use on consumer-grade hardware~\cite{deng2025vggt,maggio2025vggt}.

\begin{figure}[t]
    \captionsetup{font=small}
    \centering
    \includegraphics[width=.93\linewidth]{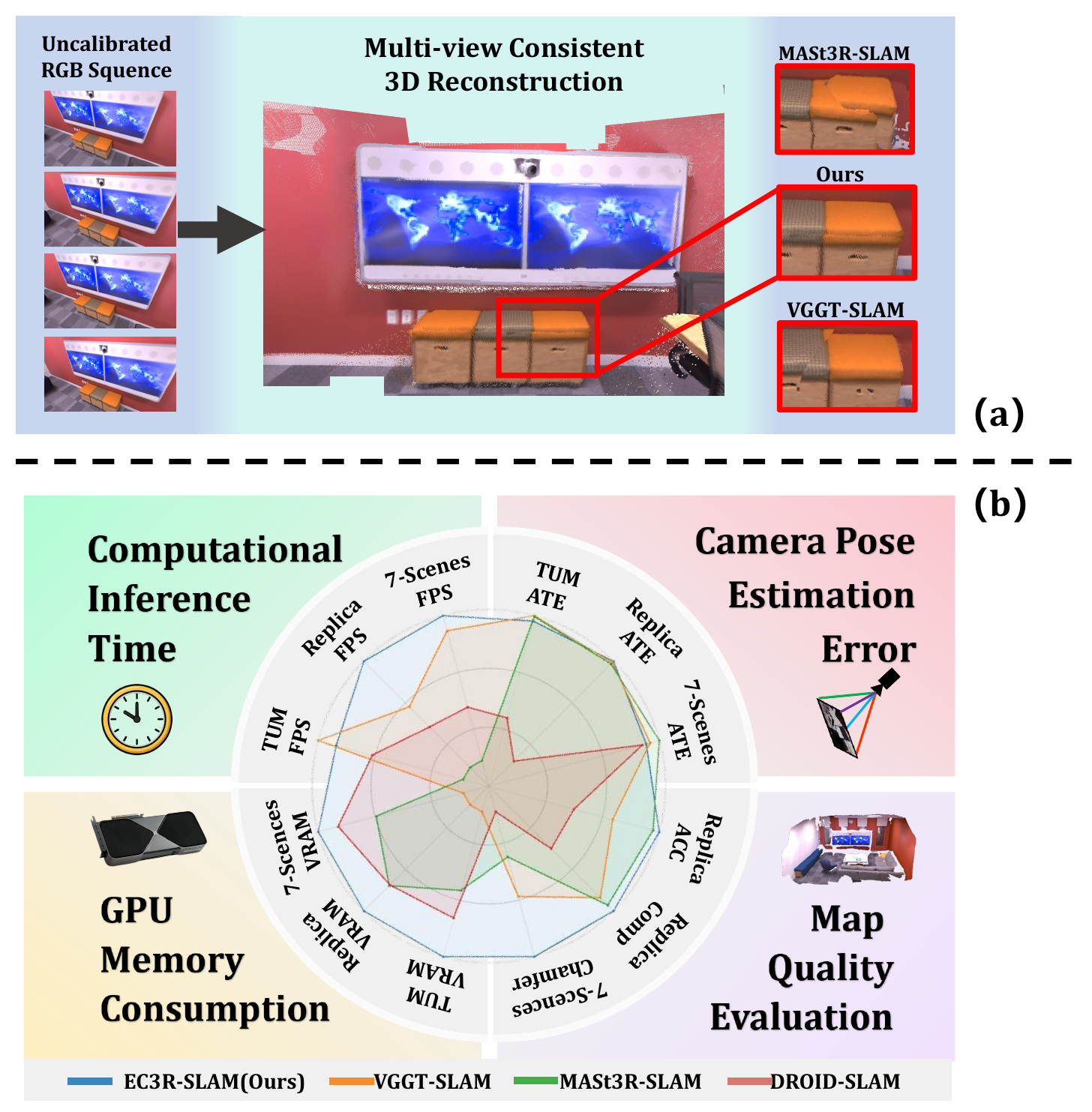}
    \captionsetup{width=\linewidth}
    \caption{(a) Our method achieves real-time multi-view consistent 3D reconstruction from uncalibrated RGB sequence.
(b) Benchmark results show fast inference and low GPU memory use with competitive accuracy, highlighting its efficiency.}
    \label{fig:eyecather}
    \vspace{-15pt}
\end{figure}

To overcome these limitations, we propose EC3R-SLAM, an \textbf{E}fficient and \textbf{C}onsistent \textbf{3}D \textbf{R}econstruction framework for calibration-free monocular dense SLAM. Our framework employs lightweight feature-based tracking to select keyframes, while only a small subset of frames (five in our implementation) is forwarded to VGGT for feed-forward reconstruction, generating local submaps that are subsequently fused into a global map. This strategy preserves the efficiency of feed-forward models while markedly reducing computational and memory demands: our pipeline operates with less than 10 GB of GPU memory and runs at more than 30 FPS. By comparison, the concurrent work VGGT–SLAM~\cite{maggio2025vggt} feeds 32 frames at once, 
whereas VGGT-Long~\cite{deng2025vggt} feeds more than 60 frames at once into VGGT, usually pushing memory usage beyond 20GB, which constrains practical deployment on resource-constrained robotic platforms. However, fusing multiple small submaps can introduce severe inconsistencies, a limitation also observed in VGGT\textendash SLAM and MASt3R\textendash SLAM~\cite{murai2024mast3r} (see Fig.~\ref{fig:eyecather}a). To mitigate this, we incorporate both local and global loop closure modules that establish mid-term and long-term associations, thereby enforcing multi-view consistency and alleviating misalignment.
As a result, EC3R-SLAM achieves accurate, real-time, and resource-efficient dense mapping and camera pose estimation, as illustrated in Fig.~\ref{fig:eyecather}b. 

\begin{figure*}[t]
    \centering
    \captionsetup{font=small}
    \includegraphics[width=\linewidth]{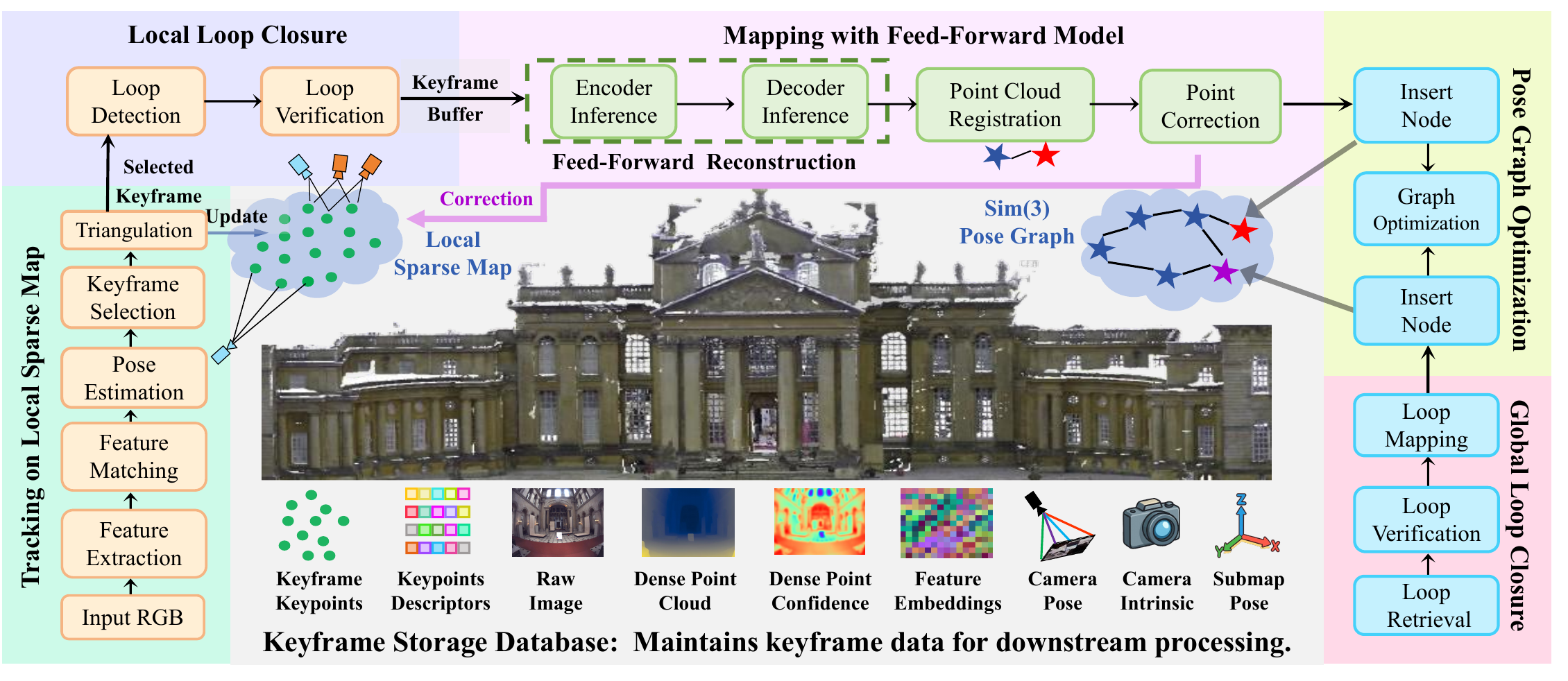}
    \caption{System overview. The RGB images are first processed in the tracking module, where keyframes are selected and used for local loop closure to identify similar frames. 
The verified keyframes are stored in the keyframe buffer, and once a sufficient number of keyframes are accumulated, they are passed to the mapping module to generate reconstruction information, which is stored in the database. 
At the same time, the global loop closure module retrieves features from the database for loop detection and performs pose graph optimization. }
    \label{fig:Pipeline}
    \vspace{-15pt}
\end{figure*}
\noindent In summary, our main contributions are:
\begin{enumerate}
    \item We present, for the first time, an innovative method that combines lightweight feature point-based tracking with feed-forward 3D reconstruction model-based mapping, enabling an efficient pipeline for localization and mapping.
    \item We propose a novel data association strategy that integrates local loop and global loop closure, and injects loop information into pose graph optimization, thereby enhancing multi-view consistency and significantly improving the overall accuracy of the system.
    \item By unifying these components, we propose a new uncalibrated monocular dense SLAM framework that delivers competitive results across multiple benchmarks, while maintaining low GPU memory usage and real-time performance.

\end{enumerate}
\section{Related Works}
\label{sec:related_work}
We first review recent works on how decoupled tracking and mapping can improve the efficiency of SLAM. We then introduce approaches that leverage data association to ensure map consistency. Finally, we review recent developments in 3D reconstruction models.
\subsection{Decoupled Tracking and Mapping}  
To improve efficiency, many SLAM systems adopt a decoupled design in which tracking and mapping are handled separately. 
The frontend typically employs lightweight methods for camera tracking and keyframe selection~\cite{mur2015orb,ORBSLAM3_TRO,teed2021droid}, 
while the backend focuses on constructing dense maps. 
For instance, Orbeez-SLAM~\cite{chung2023orbeez} leverages ORB-SLAM2~\cite{mur2015orb} for tracking to construct an implicit neural map, while other methods~\cite{zhang2024hi,sandstrom2025splat,zhang2024glorie,zhang2023go} employ DROID-SLAM~\cite{teed2021droid} to enable dense mapping.
However, the loose coupling between tracking and mapping in these systems often leads to suboptimal information usage and increased computational overhead. 
In contrast, our framework tightly couples an ultra-lightweight tracking module with mapping, substantially improving both efficiency and accuracy.
\subsection{Mid-Term and Long-Term Data Association} Mid-term data association links frames captured from nearby camera positions. For example, ORB-SLAM achieves this by projecting map points into the estimated camera pose. 
Long-term data association, conversely, relies on place recognition techniques, 
which can be implemented using either traditional bag-of-words models~\cite{DBoW3} 
or learning-based approaches such as NetVLAD~\cite{arandjelovic2016netvlad} and SALAD~\cite{izquierdo2024optimal}.
While many existing frameworks employ only one type of data association~\cite{matsuki2024gaussian,murai2024mast3r,maggio2025vggt,teed2021droid}, our approach integrates both. Moreover, long-term associations are directly obtained from the embeddings of our feed-forward reconstruction model, 
eliminating the need for extra place recognition networks. 
This design further improves accuracy while preserving real-time efficiency.

\subsection{Uncalibrated reconstruction}
DUSt3R~\cite{wang2024dust3r} and MASt3R~\cite{duisterhof2024mast3r} pioneered this direction, but multi-view reconstruction still required time-consuming post-processing. Subsequent methods such as Spann3R~\cite{wang2024spann3r}, SLAM3R~\cite{slam3r}, CUT3R~\cite{wang2025continuous}, and MASt3R-SLAM~\cite{murai2024mast3r} extended these to sequence-based reconstruction. While these approaches substantially improve real-time capability, they rely on computationally expensive dense neural inference, as the network must process every frame. VGGT-SLAM mitigates this by selecting keyframes via optical flow and applying a feed-forward model for reconstruction, yet it still incurs high GPU memory consumption. In our method, the tight coupling of tracking and mapping ensures that dense inference is avoided and memory consumption is minimized, thereby enabling efficient and scalable reconstruction.


\section{Method}
\autoref{fig:Pipeline} provides an overview of our system architecture.
In the following sections, we detail each component of our framework, including tracking (\ref{sec:tracking}), local loop closure (\ref{sec:local_loop}), mapping (\ref{sec:Mapping}), global loop closure (\ref{sec:global_loop}), and pose graph optimization (\ref{sec:pgo_optimization}).

\subsection{Tracking on the Local Sparse Map}
~\label{sec:tracking}
In this stage, we extract and match visual features across frames, perform pose estimation, and select the keyframes.  Unlike traditional approaches, these operations are carried out on the local sparse map, which serves as the basis for maintaining robust and efficient tracking.

\subsubsection{Local Sparse Map} ~\label{subsec:local_sparse_map}
Tracking in our system relies on a local sparse map, which stores the 3D points corresponding to the keypoints of the current keyframe. 
These points are updated whenever a new keyframe is selected and are continuously corrected during the mapping process.
\subsubsection{Initialization} ~\label{subsec:Initialization}
Initially, a set of selected $N$ frames is forwarded to the mapping module to estimate the camera intrinsics and perform an initial 3D reconstruction, resulting in the creation of an initial local sparse map.
\begin{figure}[t]
    \centering
    \includegraphics[width=0.9\linewidth]{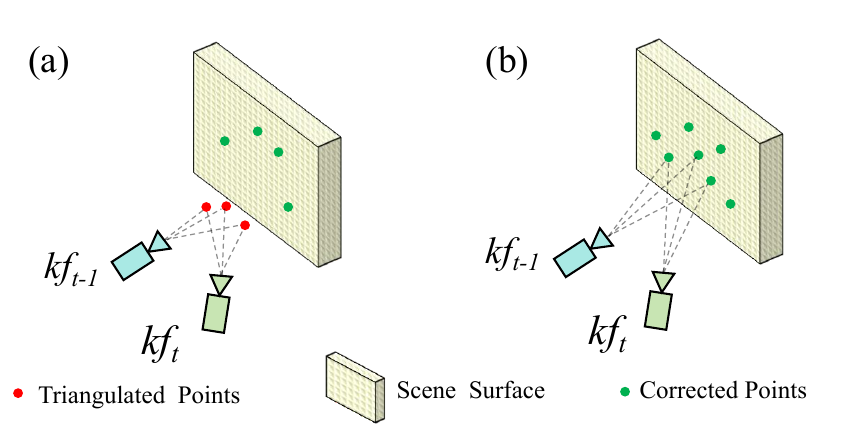}
    \caption{Illustration of point correction. 
(a) Before correction. (b) After point correction .
}
    \label{fig:point_correction}
    \vspace{-15pt}
\end{figure}
\subsubsection{Per-Frame Tracking}~
\label{subsec:Per_Frame_Tracking}
For each incoming frame, we employ XFeat~\cite{potje2024xfeat}, an efficient learning-based feature matching network, to extract keypoints and their descriptors. 
The descriptors of the current frame are matched with 3D points in the local sparse map $\{\mathbf{X}_i\}_{i=1}^N$, yielding 2D--3D correspondences $\{(\mathbf{x}_i,\mathbf{X}_i)\}_{i=1}^N$.
Given these correspondences, the camera pose $(\mathbf{R}, \mathbf{t})$ of the current frame is estimated by solving a Perspective-$n$-Point (PnP) problem that minimizes the reprojection error:
\begin{equation}
\min_{\mathbf{R}, \mathbf{t}} \sum_{i=1}^{N} \left\| \pi\!\left(\mathbf{R}\mathbf{X}_i + \mathbf{t}\right) - \mathbf{x}_i \right\|^2,
\label{eq:pnp}
\end{equation}
where $\pi(\cdot)$ denotes the perspective projection function.
To ensure robustness to outliers in the correspondences, we solve the problem using a RANSAC-based PnP algorithm~\cite{fischler1981random,lepetit2009ep}, from which the inlier ratio $\tau_{\text{inlier}}$ is computed.

\subsubsection{Keyframe Selection and Triangulation}~
\label{subsec:kf_selection_triangulation}
If the inlier ratio $\tau_{\text{inlier}}$ falls below a threshold $\tau_1$, or if pose estimation fails, indicating degraded tracking quality, the current frame is promoted to a new keyframe. 
Once selected as a keyframe, triangulation is performed to add new points into the local sparse map.
\subsection{Local Loop Closure for Mid-Term Data Association}
\label{sec:local_loop}
To enforce mid-term data association, we integrate a local loop closure mechanism immediately after the tracking stage, whose main purpose is to associate the current keyframe with spatially nearby frames.

\subsubsection{Loop Detection}~
Potential local loop frames are identified by projecting the current local sparse map onto a set of temporally adjacent keyframes, as illustrated in Fig.~\ref{fig:Pipeline}. 
This set is defined as
\[
\left\{ \text{KF}_k \right\}_{k = i - r_{\text{local}}}^{i + r_{\text{local}}}, \quad \text{where } i + r_{\text{local}} = n - 2N_{\text{map}},
\]
with $n$ denoting the index of the current keyframe. 
A keyframe within this set is regarded as a loop closure candidate if the number of 3D points from the local sparse map projected onto its image plane exceeds a predefined threshold $\tau_p$.
\subsubsection{Loop Verification}~
\label{subsec:loop_verification}
To verify loop-closure candidates, we compute the inlier ratio $r_{\text{inlier}}$ using RANSAC homography estimation. 
If $r_{\text{inlier}} \leq \tau_2$, the candidate is considered dissimilar and discarded. 
If $r_{\text{inlier}} > \tau_2$, the candidate is appended to the keyframe buffer for further processing. 
In particular, when $r_{\text{inlier}} > \tau_1$, the candidate is regarded as highly similar, and the current keyframe is replaced by the loop candidate.

\subsubsection{Keyframe Buffer}~
\label{subsec:kf_buffer}
The keyframe buffer consists of two components:  
(1) new keyframes obtained from the tracking module, and  
(2) old frames, including loop closure candidates and the most recent keyframe from the previous buffer.  
Once the number of buffered frames exceeds the predefined threshold $N$,  
the buffer is flushed, and all accumulated keyframes are forwarded to the mapping module for dense reconstruction.  
After flushing, the buffer is re-initialized with the latest keyframe to ensure continuity for subsequent processing.
\subsection{Feed-Forward Model Reconstruction}
\label{sec:Mapping}
This module runs in a separate process parallel to tracking and enables the system to construct a dense point cloud of the scene. 
We adopt VGGT~\cite{wang2025vggt} to infer a local submap, which is subsequently aligned with the global map.

\subsubsection{Feed-Forward Inference}~
\label{subsec:feedforward}
To improve efficiency, encoder inference is only performed on new keyframes from the keyframe buffer. 
Given the RGB image $\mathbf{I}_1$ of a new keyframe, the image encoder $\mathcal{E}(\cdot)$  produces image feature embeddings:
\begin{equation}
\mathbf{E}_{1} = \mathcal{E}(\mathbf{I}_{1}),
\end{equation}
where $\mathbf{E}_1 = \{ \mathbf{e}_1, \cdots, \mathbf{e}_{k}\}$ denotes the keyframe feature embeddings. 
For old keyframes, their embeddings $\mathbf{E}_2 = \{ \mathbf{e}_{k+1}, \cdots, \mathbf{e}_{n}\}$ are directly retrieved from the keyframe storage database.

The embeddings $\mathbf{E}_1$ and $\mathbf{E}_2$ are concatenated and 
fed into the prediction decoder $\mathcal{D}(\cdot)$, which in the VGGT 
architecture refers to the remaining part of the network after the image encoder:
\begin{equation}
\mathbf{D}, \mathbf{C}, \mathbf{g} = \mathcal{D} \left( 
\text{Concat} \left( \mathbf{E}_{1}, \mathbf{E}_{2} \right) 
\right),
\end{equation}
where $\mathbf{D}$ denotes the predicted depth maps, 
$\mathbf{C}$ the corresponding confidence maps that quantify the reliability 
of each depth estimate, and $\mathbf{g}$ the camera parameters 
(intrinsics and extrinsics).
Following~\cite{maggio2025vggt,wang2025vggt}, the current submap is obtained by inverse-projecting the estimated depths $\mathbf{D}$ using the projection matrices derived from $\mathbf{g}$. 
This submap consists of a dense point cloud, where each pixel in the keyframe corresponds to a 3D point with an associated confidence score. 
The submap is defined with respect to the coordinate frame of the first camera.

\subsubsection{Sim3 Point Cloud Registration}~
\label{subsec:sim3_pcr}
Current submap contains a subset of 3D points 
$\mathbf{P} = \{\mathbf{p}_i\}_{i=1}^{N}$ 
originating from old keyframes. 
These points can be associated with their counterparts 
$\mathbf{Q} = \{\mathbf{q}_i\}_{i=1}^{N}$ 
in other submaps that belong to the same old keyframes, thereby forming reliable 3D--3D correspondences. 
Their corresponding confidence scores are normalized into weights 
$\mathbf{w} = \{w_i\}_{i=1}^{N}$. 
We then estimate the optimal Sim(3) transformation 
$\mathbf{S} = \{s, \mathbf{R}, \mathbf{t}\} \in \text{Sim}(3)$ 
by minimizing the weighted alignment error:
\begin{equation}
\min_{s, \mathbf{R}, \mathbf{t}} \sum_{i=1}^{N} w_i
\left\| s \mathbf{R}\,\mathbf{p}_i + \mathbf{t} - \mathbf{q}_i \right\|^2,
\label{eq:sim3}
\end{equation}
where $s \in \mathbb{R}^+$ is a scale factor, $\mathbf{R} \in \text{SO}(3)$ a rotation, and $\mathbf{t} \in \mathbb{R}^3$ a translation. 
We solve \eqref{eq:sim3} using a weighted extension of Umeyama’s closed-form Sim(3) algorithm~\cite{umeyama1991least}. 
This step yields the transformation between the current submap and other connected submaps, 
which is then incorporated into the pose graph as a relative Sim(3) constraint.

\begin{figure}[t]
    \centering
    \includegraphics[width=0.8\linewidth]{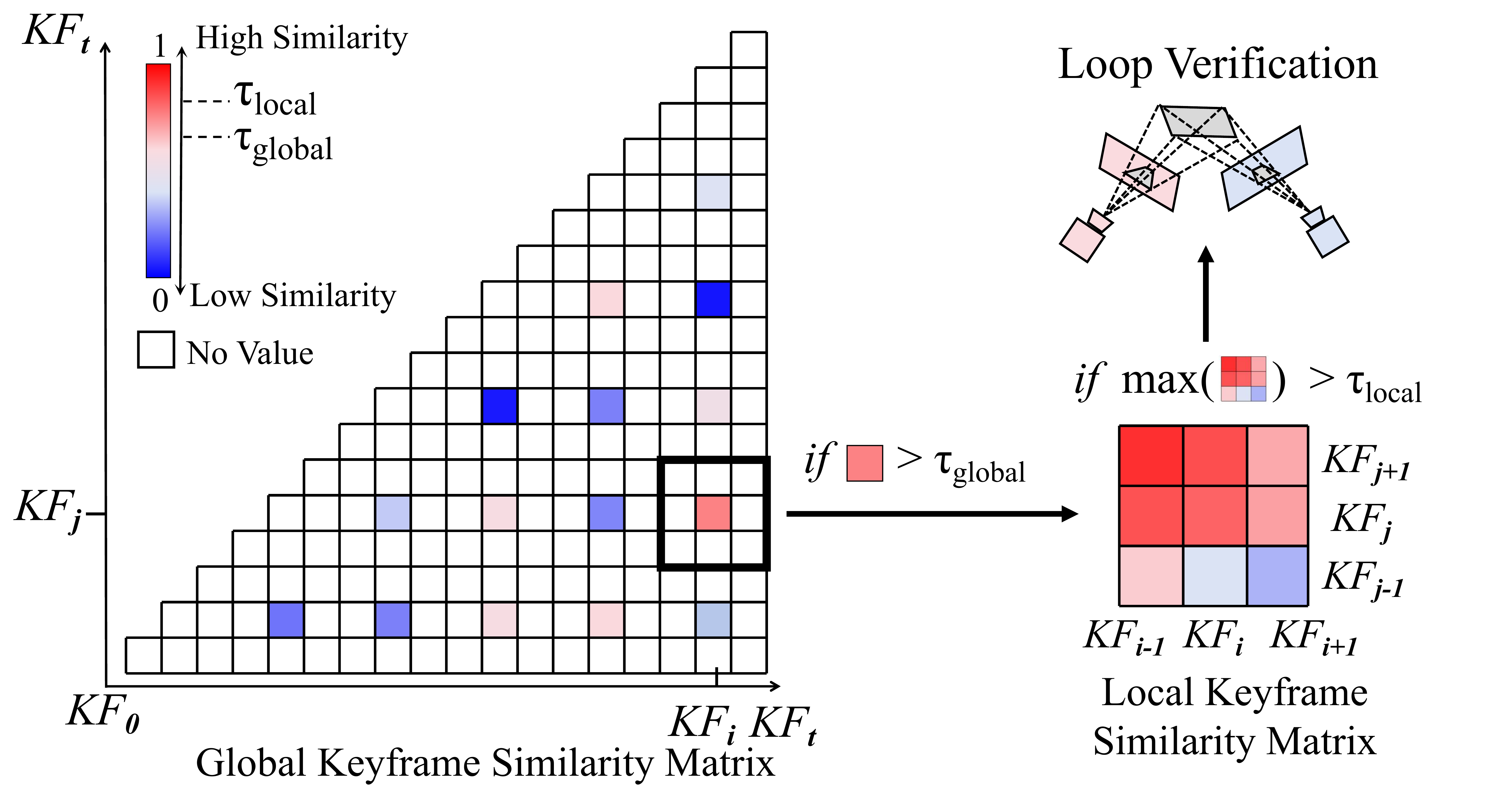}
  \caption{\small We perform loop detection by computing a similarity matrix and filtering it with global and local thresholds, followed by homography-based verification.}
    \label{fig:global_loop_closure}
    \vspace{-15pt}
\end{figure}
\subsubsection{Point Correction}~
\label{subsec:point_correction}
Through the estimated transformations between submaps, we obtain the global coordinates of all points in the current submap, which are regarded as accurate. 
A subset of these points has correspondences with those in the local sparse map. 
For such points, we replace their coordinates with the accurate global ones, as illustrated in Fig.~\ref{fig:point_correction}, thereby reducing the accumulated drift of tracking. 
Moreover, to further prevent the frontend from being lost, whenever the number of frames in the keyframe buffer reaches 2, we trigger an additional mapping operation. 
This operation is solely used to obtain the current keyframe's points relative to the global frame, which are then applied for correction.

\begin{figure*}[t]
    \centering
\includegraphics[width=\textwidth]{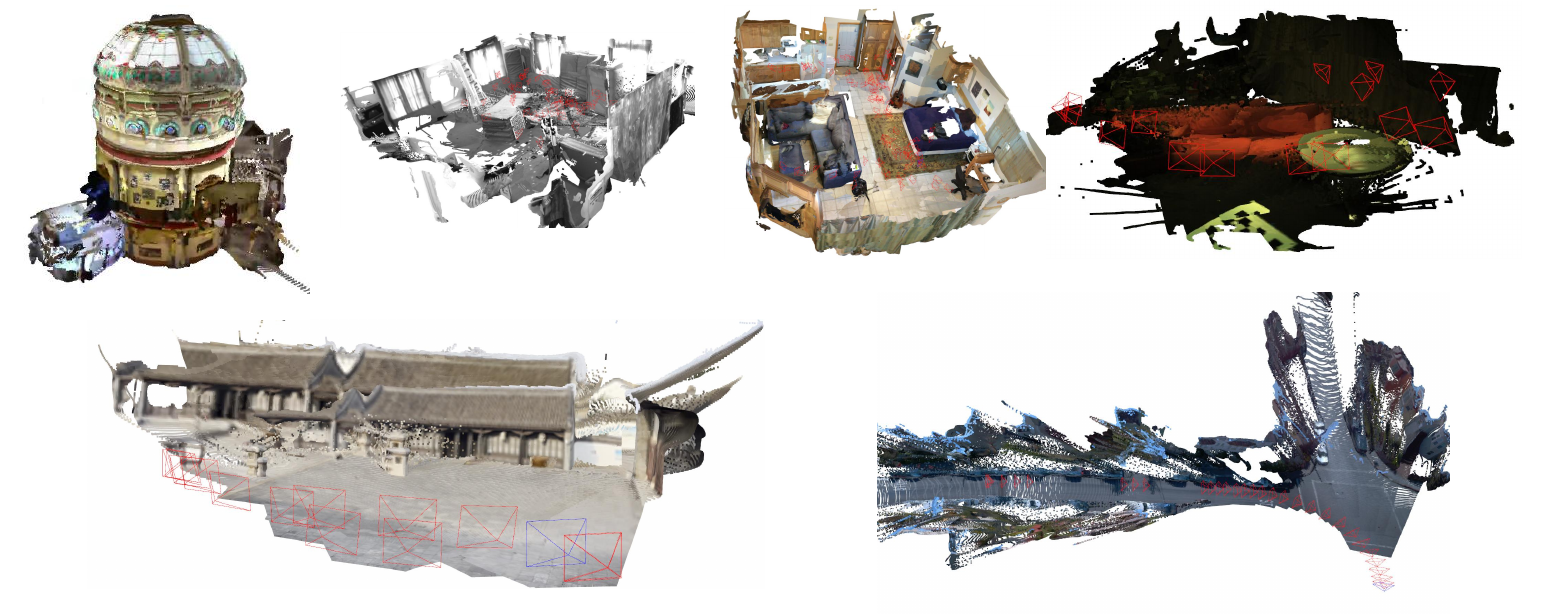}
    \caption{EC3R-SLAM can generalize to new datasets.We show results from  Tanks and Temples~\cite{knapitsch2017tanks}, ScanNet~\cite{dai2017scannet}, EuRoC~\cite{burri2016euroc}, Waymo open~\cite{sun2020scalability}, ETH3D~\cite{schops2019bad},and DL3DV~\cite{dl3dv}.}
    \label{fig:generalization}
\end{figure*}

{\renewcommand{\arraystretch}{0.95}
\begin{table*}[ht]
\centering
\setlength{\tabcolsep}{4.8pt}
\scriptsize 
\caption{Evaluation of dense monocular SLAM methods on the TUM-RGBD dataset under the uncalibrated setting.}
\label{tab:tumrgbd}
\resizebox{\textwidth}{!}{ 
\begin{tabular}{llcccccccccc|cc}
\hline
\multirow{2}{*}{\textbf{Map type}} &
\multirow{2}{*}{\textbf{Method}} &
\multicolumn{9}{c}{\textbf{Sequence}} &
\multirow{2}{*}{\textbf{Avg$\downarrow$}} &
\multirow{2}{*}{\textbf{VRAM$\downarrow$}} &
\multirow{2}{*}{\textbf{FPS$\uparrow$}} \\
\cline{3-11}
& & \textbf{360} & \textbf{desk} & \textbf{desk2} & \textbf{floor} & \textbf{plant} & \textbf{room} & \textbf{rpy} & \textbf{teddy} & \textbf{xyz} & & & \\
\hline
\multirow{2}{*}{NeRF}
& GlORIE-SLAM   & 0.194 & 0.028 & 0.105 & 0.062 & 0.036 & 0.871 & 0.051 & \underline{0.042} & \underline{0.014} & 0.155 & 18.9 & <1 \\
& GO-SLAM       & 0.195 & 0.031 & 0.198 & 0.077 & 0.049 & 0.865 & 0.058 & 0.049 & \underline{0.014} & 0.171 & 20.8 & 7 \\
\hline
\multirow{4}{*}{3DGS}
& Photo-SLAM    & 0.165 & 0.028 & 0.880 & \textbf{0.055} & 0.705 & 0.924 & \textbf{0.027} & 0.933 & \textbf{0.013} & 0.414 & \underline{7.6} & 23 \\
& Mono-GS       & 0.159 & 0.043 & 0.689 & 0.603 & 0.616 & 0.726 & 0.041 & 0.100 & 0.022 & 0.333 & 13.0 & 2 \\
& Splat-SLAM    & 0.205 & \underline{0.026} & 0.101 & 0.062 & 0.038 & 0.879 & 0.051 & \underline{0.042} & \underline{0.014} & 0.158 & 18.8 & 1 \\
& Hi-SLAM2      & 0.223 & 0.068 & 12.62 & 0.305 & 0.059 & 2.204 & 0.035 & 0.213 & 0.033 & 1.750 & 22.1 & 3 \\
\hline
\multirow{5}{*}{Point Cloud}
& DROID-SLAM    & 0.194 & 0.034 & 0.822 & 0.168 & 0.038 & 0.975 & 0.056 & 0.059 & 0.019 & 0.263 & 13.1 & 23 \\
& MASt3R-SLAM   & \underline{0.070} & 0.035 & 0.055 & \underline{0.056} & \underline{0.035} & \underline{0.119} & 0.041 & 0.115 & 0.019 & \underline{0.061} & 15.8 & 13 \\
& VGGT-SLAM     & \textbf{0.064} & \textbf{0.024} & \textbf{0.036} & 0.126 & \textbf{0.023} & 0.163 & \underline{0.032} & \textbf{0.037} & 0.017 & \textbf{0.059} & 23.5 & \textbf{34} \\
& Ours(w.Fast3R)& 0.129 & 0.056 & 0.091 & 0.067 & 0.113 & 0.145 & 0.052 & 0.143 & 0.022 & 0.091 & \textbf{7.1} & 27 \\
& Ours          & 0.101 & 0.038 & \underline{0.050} & \textbf{0.055} & 0.097 & \textbf{0.101} & 0.045 & 0.125 & 0.018 & 0.070 & 9.3 & \underline{31} \\
\hline
\end{tabular}
}
\end{table*}
}

\subsubsection{Keyframe Storage Database}~
\label{subsec:kf_database}
As illustrated in Fig.~\ref{fig:Pipeline}, the keyframe storage database maintains all keyframe information obtained during the tracking and mapping stages. 
This information is later exploited for loop detection and loop retrieval. 
To save computational resources, the database is stored on the CPU.
\begin{figure*}[t]
  \centering
  \includegraphics[width=\textwidth]{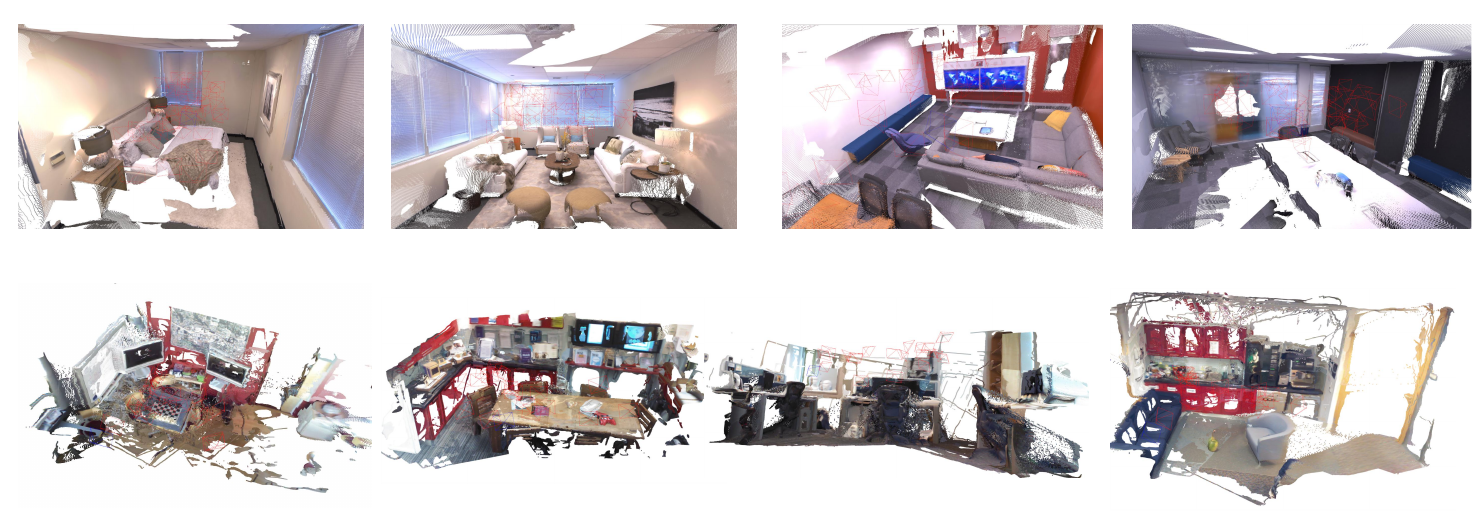}
  \caption{Qualitative reconstruction results on the Replica and 7-Scenes datasets. }
\label{fig:qual_replica_7scenes}
\end{figure*}
\subsection{Global Loop Closure for Long-Term Data Association}
\label{sec:global_loop}
We employ a novel approach for global loop closure. 
This module runs in a separate thread and leverages the keyframe features extracted in Section~\ref{sec:Mapping} to detect global loop closures, thereby establishing long-term associations in the global map. 
An overview of the pipeline is presented in Fig.~\ref{fig:global_loop_closure}.

\subsubsection{Loop Retrieval and Verification}~
 We first construct a sparse similarity matrix by measuring feature similarity between every $N$-th keyframe stored in the database. 
If two keyframes are sufficiently similar, their neighboring frames are also considered and incorporated into the similarity matrix. 
When the maximum similarity score exceeds the threshold $\tau_{\text{local}}$, we further verify the candidate pair by estimating a homography and counting the inliers ratio. 
If this verification succeeds, a loop closure is confirmed.
\subsubsection{Loop Mapping}~
We select the embeddings of the top-$N$ most similar keyframes from the local similarity matrix. 
These embeddings are passed through the decoder inference of the mapping module (Section~\ref{sec:Mapping}) to generate a submap, 
which is subsequently registered and inserted into the pose graph.
\subsection{Pose Graph Optimization}
\label{sec:pgo_optimization}
The relative Sim(3) constraints $T_{ij}$ obtained in Section~\ref{subsec:sim3_pcr} are incorporated as edges into a submap-level pose graph, 
where each node represents the global pose of a submap $T_t \in \mathrm{Sim}(3)$ in the world coordinate frame. 
The goal of pose graph optimization is to jointly refine all submap poses by enforcing consistency across the 3D--3D correspondences.

During optimization, each transformation $T$ is mapped to a minimal representation in $\mathbb{R}^7$ of the associated Lie algebra using the logarithmic mapping function $\log_{\mathrm{Sim}(3)}(\cdot)$. 
Given the constructed pose graph, the error on an edge $(i,j)$ is defined as
\begin{equation}
e_{i,j} = \log_{\mathrm{Sim}(3)} \!\left( T_{ij}^{-1} T_i^{-1} T_j \right).
\end{equation}

The overall objective is to minimize the total energy
\begin{equation}
\chi^2(T_1,\dots,T_m) 
= \sum_{(i,j)\in\mathcal{E}} e_{i,j}^\top \Omega_{i,j}\, e_{i,j},
\end{equation}
where $\Omega_{i,j}$ denotes the information matrix associated with the measurement. 
The absolute submap poses $\{T_t\}$ are initialized by chaining the relative Sim(3) transformations, 
and the optimization is carried out using the Levenberg--Marquardt algorithm implemented in \textit{PyPose}~\cite{wang2023pypose}, 
which enables efficient Lie-group optimization on $\mathrm{Sim}(3)$. 
This process yields globally consistent submap poses and effectively reduces accumulated drift.

{\renewcommand{\arraystretch}{0.95}
\begin{table*}[ht]
\centering
\setlength{\tabcolsep}{4pt}
\scriptsize 
\caption{Evaluation of  dense monocular SLAM methods on the Replica dataset  under the uncalibrated setting.}
\label{tab:replica}
\resizebox{\textwidth}{!}{ 
\begin{tabular}{llccccccccc|cc}
\hline
\multirow{2}{*}{\textbf{Map type}} &
\multirow{2}{*}{\textbf{Method}} &
\multicolumn{8}{c}{\textbf{Sequence}} &
\multirow{2}{*}{\textbf{Avg$\downarrow$}} &
\multirow{2}{*}{\textbf{VRAM$\downarrow$}} &
\multirow{2}{*}{\textbf{FPS$\uparrow$}} \\
\cline{3-10}
& & \textbf{Room0} & \textbf{Room1} & \textbf{Room2} & \textbf{Office0} & \textbf{Office1} & \textbf{Office2} & \textbf{Office3} & \textbf{Office4} & & & \\
\hline
\multirow{2}{*}{NeRF}
& GlORIE-SLAM & 0.374 & 0.119 & 0.117 & 0.337 & 0.419 & 0.050 & 0.472 & 0.282 & 0.234 & 17.9 & <1 \\
& GO-SLAM     & 0.378 & 0.118 & 0.098 & 0.033 & 0.416 & 0.059 & 0.543 & 0.323 & 0.246 & 16.2 & 12 \\
\hline
\multirow{4}{*}{3DGS}
& Photo-SLAM  & 0.363 & 0.139 & \underline{0.056} & \underline{0.030} & 0.423 & 0.094 & 0.313 & 0.382 & 0.222 & 9.4 & 2 \\
& Mono-GS     & 0.716 & 0.518 & 0.167 & 0.231 & 0.439 & 0.296 & 0.269 & 0.679 & 0.456 & 12.3 & 2 \\
& Splat-SLAM  & 0.374 & 0.119 & 0.107 & 0.034 & 0.418 & 0.052 & 0.472 & 0.287 & 0.233 & 22.6 & 4 \\
& Hi-SLAM2    & 0.314 & 0.113 & 0.116 & \underline{0.031} & 0.418 & \underline{0.040} & 0.253 & 0.249 & 0.192 & 15.2 & 16 \\
\hline
\multirow{5}{*}{Point Cloud}
& DROID-SLAM      & 0.313 & 0.111 & 0.125 & \textbf{0.029} & 0.421 & 0.045 & 0.253 & 0.249 & 0.193 & 13.1 & 28 \\
& MASt3R-SLAM     & \textbf{0.023} & \underline{0.035} & 0.103 & 0.038 & \underline{0.033} & 0.047 & 0.035 & \underline{0.045} & 0.045 & 13.0 & 20 \\
& VGGT-SLAM       & \underline{0.030} & 0.082 & 0.057 & 0.061 & \textbf{0.028} & \textbf{0.024} & \underline{0.028} & \textbf{0.032} & \underline{0.043} & 25.4 & \underline{36} \\
& Ours(w.Fast3R)  & 0.042 & \textbf{0.030} & 0.077 & 0.129 & 0.042 & 0.051 & 0.035 & 0.074 & 0.060 & \textbf{6.8} & 35 \\
& Ours            & 0.038 & 0.049 & \textbf{0.027} & 0.043 & 0.034 & 0.059 & \textbf{0.025} & 0.053 & \textbf{0.041} & \underline{9.1} & \textbf{45} \\
\hline
\end{tabular}
}
\vspace{-4mm}
\end{table*}
}

\begin{table}[t]
\centering
\caption{Comparison of reconstruction quality and camera pose accuracy on the 7-Scenes dataset.}
\label{tab:7scenes_recon}
\scriptsize
\renewcommand{\arraystretch}{0.95}
\resizebox{\linewidth}{!}{
\begin{tabular}{l|ccc|ccc}
\hline
Method & Acc ↓ & Comp ↓ & Cham ↓ & ATE ↓ & VRAM ↓ & FPS ↑ \\
\hline
Spann3R@20*        & 0.069 & \underline{0.047} & 0.058 & --    & --   & --  \\
Spann3R@2*         & 0.124 & \textbf{0.043} & 0.084 & --    & --   & --  \\
SLAM3R             & \underline{0.038} & 0.070 & \underline{0.054} & 0.084 &  -- &  -- \\
DROID-SLAM         & 0.099 & 0.057 & 0.078 &0.078 & 11.2 & 24 \\
MASt3R-SLAM        & 0.065 & 0.067 & 0.056 & \textbf{0.065} & 15.2 & 17 \\
VGGT-SLAM          & 0.052 & 0.059 & 0.056 & \underline{0.072} & 23.4 & \underline{34} \\
Ours (w. Fast3R)   & 0.044 & 0.076 & 0.060 & 0.090 & \textbf{7.0} & 32 \\
Ours               & \textbf{0.025} &0.054 & \textbf{0.040} & 0.075 & \underline{9.1} & \textbf{36} \\
\hline
\end{tabular}
}

\vspace{1mm}
\raggedright
\footnotesize{\textit{@n} indicates a keyframe every \textit{n} images.\\
* indicates results reported in MASt3R-SLAM.}
\vspace{-6mm}
\end{table}

\section{Experiment}
We first introduce the experimental setup, followed by a detailed analysis of localization and mapping results, runtime, GPU memory consumption, and ablation studies.
\subsection{Experimental Setup}
\subsubsection{Implementation Details}~ 
We implement EC3R-SLAM in Python and adopt all neural network models directly from their official implementations. Unless otherwise specified, all experiments were conducted on a desktop equipped with an NVIDIA GeForce RTX 5090 GPU and an Intel Core i9-12900KF CPU. In addition, we replace VGGT with Fast3R~\cite{yang2025fast3r} across the entire system and conduct the same set of experiments, reported in the tables as \textit{Ours (w. Fast3R)}. 
{\renewcommand{\arraystretch}{0.95}
\begin{table*}[ht]
\centering
\setlength{\tabcolsep}{4.8pt}
\scriptsize 
\caption{ATE RMSE comparison of uncalibrated dense monocular SLAM methods on the 7-Scenes dataset. 
Lower is better for ATE/VRAM; higher is better for FPS. Bold = best, underline = second best.}
\label{tab:7scenesate}
\resizebox{\textwidth}{!}{
\begin{tabular}{lccccccc|ccc}
\hline
\multirow{2}{*}{\textbf{Method}} &
\multicolumn{7}{c}{\textbf{Sequence}} &
\multirow{2}{*}{\textbf{Avg$\downarrow$}} &
\multirow{2}{*}{\textbf{VRAM$\downarrow$}} &
\multirow{2}{*}{\textbf{FPS$\uparrow$}} \\
\cline{2-8}
& \textbf{Chess} & \textbf{Fire} & \textbf{Heads} & \textbf{Office} & \textbf{Pump.} & \textbf{Kitc.} & \textbf{Stairs} &  &  & \\
\hline
GlORIE\mbox{-}SLAM     & 0.050 & 0.043 & 0.097 & 0.187 & 0.157 & 0.100 & \underline{0.021} & 0.094 & 18.5 & 0.1 \\
GO\mbox{-}SLAM         & 0.056 & 0.040 & 0.095 & 0.210 & 0.162 & 0.107 & 0.025 & 0.099 & 13.6 & 16 \\
Photo\mbox{-}SLAM      & \textbf{0.016} & \textbf{0.021} & 1.102 & \textbf{0.075} & \textbf{0.048} & \textbf{0.013} & 0.061 & 0.191 & \textbf{5.5} & 30 \\
MonoGS                 & 0.057 & \underline{0.032} & 0.303 & 0.275 & 0.377 & 0.453 & 0.187 & 0.241 & \underline{6.1} & 2 \\
Splat\mbox{-}SLAM      & 0.050 & 0.043 & 0.095 & 0.188 & 0.157 & 0.099 & \textbf{0.020} & \underline{0.093} & 15.2 & 4 \\
Hi\mbox{-}SLAM2        & 0.043 & 0.041 & 0.267 & 0.134 & \underline{0.104} & 0.114 & 0.023 & 0.104 & 22.4 & 5 \\
DROID\mbox{-}SLAM      & 0.047 & 0.038 & 0.034 & 0.136 & 0.166 & 0.080 & 0.044 & 0.078 & 11.2 & 24 \\
MASt3R\mbox{-}SLAM     & 0.063 & 0.046 & \underline{0.029} & \underline{0.103} & 0.109 & 0.074 & 0.032 & \textbf{0.065} & 15.2 & 17 \\
VGGT\mbox{-}SLAM       & \underline{0.036} & \underline{0.026} & \textbf{0.018} & 0.105 & 0.166 & \underline{0.061} & 0.092 & \underline{0.072} & 24.3 & \textbf{38} \\
Ours (w.\ F3R)         & 0.077 & 0.059 & 0.042 & 0.108 & 0.123 & 0.081 & 0.143 & 0.090 & 7.0 & 30 \\
Ours                   & 0.050 & 0.042 & 0.059 & 0.113 & 0.143 & 0.065 & 0.050 & 0.075 & 11.4 & \underline{36} \\
\hline
\end{tabular}
}
\end{table*}
}

\subsubsection{Hyperparameters}~ We set the thresholds as follows: $\tau_{1}=0.4$ and $\tau_{2}=0.3$ for loop verification and keyframe selection, $\tau_{p}=0.7$ for local loop detection, $\tau_{\mathrm{global}}=0.93$ and $\tau_{\mathrm{local}}=0.96$ for global loop retrival, and $N=5$ for the keyframe buffer.

\begin{figure*}[ht]
  \centering
  \includegraphics[width=\textwidth]{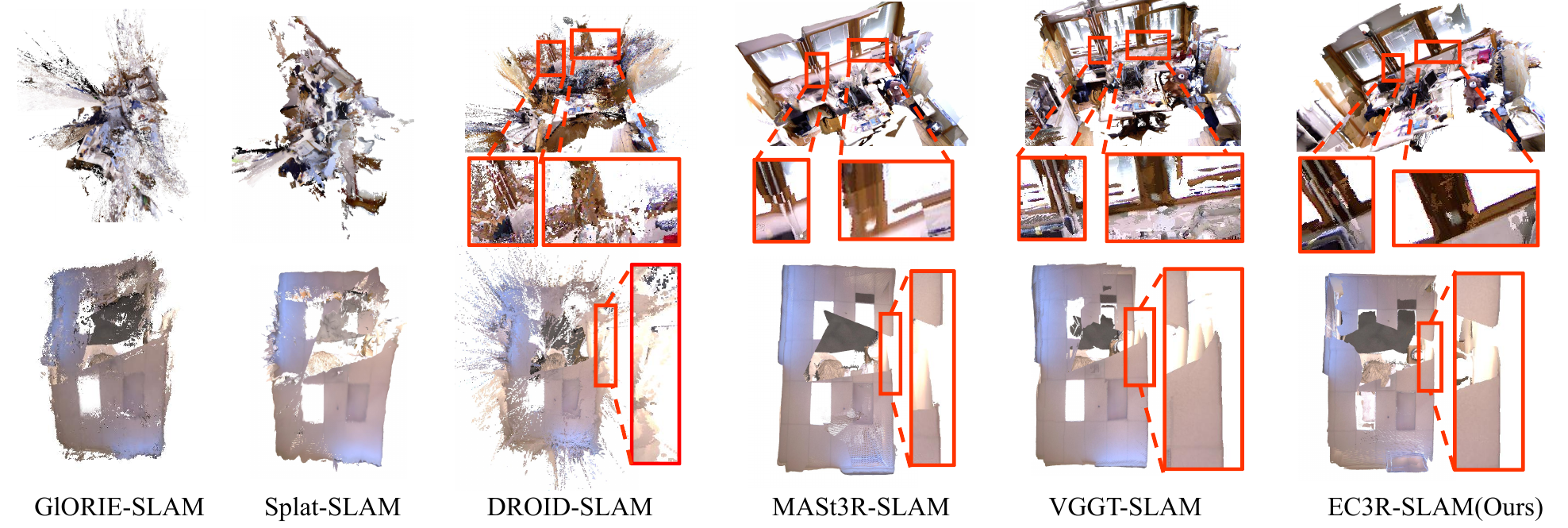}
  \caption{Qualitative comparison on TUM-RGBD fr1/room (top) and Replica Room-1 (bottom).
  Our method achieves high-quality reconstruction with both local detail preservation and global structural consistency.}
  \label{fig:qual_replica_tum}
\end{figure*}
{\renewcommand{\arraystretch}{0.95}
\begin{table*}[ht]
\centering
\caption{Comparison of 3D reconstruction accuracy and completeness (in cm) on Replica dataset.}
\label{tab:replica_results}
\setlength{\arrayrulewidth}{0.5pt}
\resizebox{\textwidth}{!}{
\begin{tabular}{c|cc|cc|cc|cc|cc|cc|cc|cc|cc}
\hline
\multirow{2}{*}{Method} & \multicolumn{2}{c|}{Room 0} & \multicolumn{2}{c|}{Room 1} & \multicolumn{2}{c|}{Room 2} & \multicolumn{2}{c|}{Office 0} & \multicolumn{2}{c|}{Office 1} & \multicolumn{2}{c|}{Office 2} & \multicolumn{2}{c|}{Office 3} & \multicolumn{2}{c|}{Office 4} & \multicolumn{2}{c}{Average} \\
\cline{2-19}
& Acc. & Comp. & Acc. & Comp. & Acc. & Comp. & Acc. & Comp. & Acc. & Comp. & Acc. & Comp. & Acc. & Comp. & Acc. & Comp. & Acc. & Comp. \\
\hline
DUSt3R*      & 3.47 & 2.50 & \textbf{2.53} & \underline{1.86} & 2.95 & 1.76 & 4.92 & 3.51 & \underline{3.09} & \underline{2.21} & 4.01 & 3.10 & 3.27 & \underline{2.25} & 3.66 & 2.61 & 3.49 & 2.48 \\
MASt3R*      & 4.01 & 4.10 & 3.61 & 3.25 & 3.13 & 2.15 & \textbf{2.57} & \textbf{1.63} & 12.85 & 8.13 & 3.13 & \underline{1.99} & 4.67 & 3.15 & 3.69 & 2.47 & 4.71 & 3.36 \\
SLAM3R       & 4.31 & 3.51 & 2.72 & 1.91 & 3.76 & 2.73 & 4.07 & 2.37 & 3.57 & 2.44 & 3.62 & 2.66 & 4.44 & 3.08 & \textbf{3.01} & 2.29 & 3.69 & 2.62 \\
Spann3R*     & 9.75 & 12.94 & 15.51 & 12.94 & 7.28 & 8.50 & 5.46 & 18.75 & 5.24 & 16.64 & 9.33 & 11.80 & 16.00 & 9.03 & 13.97 & 16.02 & 10.32 & 13.33 \\
\hline
Hi-SLAM2     & 77.78 & 34.73 & 21.83 & 30.33 & 64.64 & 26.39 & 5.74 & 6.87 & 40.82 & 96.74 & 5.47 & 7.09 & 39.36 & 22.59 & 15.37 & 10.69 & 41.58 & 31.97 \\
Splat-SLAM   & 61.67 & 25.30 & 23.56 & 31.82 & 59.33 & 34.32 & 4.52 & 4.29 & 60.71 & 80.21 & 8.12 & 6.80 & 52.40 & 22.55 & 18.80 & 10.23 & 36.99 & 29.95 \\
GlORIE-SLAM  & 68.92 & 15.22 & 29.16 & 26.69 & 60.27 & 21.66 & 4.61 & 3.69 & 21.55 & 116.11 & 11.71 & 6.10 & 49.78 & 15.03 & 17.39 & 7.74 & 39.49 & 24.39 \\
DROID-SLAM   & 33.89 & 14.85 & 23.64 & 9.37 & 30.25 & 8.01 & 5.52 & 1.96 & 38.12 & 39.61 & 5.15 & 2.65 & 24.86 & 7.59 & 17.97 & 5.23 & 22.43 & 11.16 \\
MASt3R-SLAM  & \underline{2.55} & \underline{2.01} & \underline{2.66} & \textbf{1.83} & \underline{2.14} & \underline{1.58} & \underline{2.88} & \underline{1.87} & 3.45 & 2.57 & \underline{2.80} & 2.06 & 3.75 & 2.76 & \underline{3.13} & \textbf{2.21} & \underline{2.92} & \underline{2.25} \\
VGGT-SLAM    & 3.39 & 2.56 & 7.59 & 3.37 & 2.31 & 1.71 & 5.50 & 2.43 & 22.89 & 15.39 & \textbf{2.61} & \textbf{1.79} & \underline{2.86} & 2.56 & 3.44 & \underline{2.25} & 6.32 & 4.01 \\
Ours(w.Fast3R) & 4.34 & 2.83 & 6.43 & 4.41 & 4.08 & 2.44 & 16.20 & 6.85 & 5.30 & 4.46 & 3.90 & 2.52 & 3.84 & 2.61 & 4.58 & 2.89 & 6.08 & 3.63 \\
Ours         & \textbf{2.07} & \textbf{1.63} & 2.90 & 2.03 & \textbf{1.93} & \textbf{1.57} & 4.06 & 2.48 & \textbf{2.17} & \textbf{1.69} & 3.41 & 2.43 & \textbf{2.51} & \textbf{2.08} & 3.52 & 2.66 & \textbf{2.82} & \textbf{2.07} \\
\hline
\end{tabular}
}

\vspace{1mm}
\raggedright
\footnotesize{
* indicates results reported in SLAM3R}
        \vspace{-4mm}

\end{table*}
}

\subsubsection{Datasets}~
To ensure comprehensive evaluation, we use sequences from three standard datasets: the fr1 series of TUM-RGBD~\cite{sturm2012benchmark}, seq-01 from 7-Scenes~\cite{shotton2013scene}, and 8 scenes from Replica~\cite{straub2019replica}. 
Replica and 7-Scenes provide reliable ground-truth reconstructions, making them suitable for evaluating accuracy and completeness, while TUM-RGBD is more challenging due to handheld motion, rolling shutter, and motion blur. 
We do not perform subsampling on the datasets; instead, all images are resized while preserving aspect ratio, with the longer side scaled to 518 pixels.

\subsection{Camera Pose Estimation Evaluation}

\subsubsection{Baselines}~
Our evaluation focuses on state-of-the-art monocular dense SLAM approaches, including point cloud-based methods such as MASt3R-SLAM~\cite{murai2024mast3r}, DROID-SLAM~\cite{teed2021droid}, and VGGT-SLAM~\cite{maggio2025vggt}, as well as NeRF~\cite{mildenhall2021nerf}-based methods (GlORIE-SLAM~\cite{zhang2024glorie}, GO-SLAM~\cite{zhang2023go}) and 3D Gaussian Splatting(3DGS)~\cite{kerbl20233d}-based methods (MonoGS~\cite{matsuki2024gaussian}, Splat-SLAM~\cite{sandstrom2025splat}, Hi-SLAM2~\cite{zhang2024hi}).
All methods, except for MASt3R-SLAM and VGGT-SLAM, rely on known camera intrinsic parameters. 
To ensure a comparison under the uncalibrated setting, we follow the self-calibration protocol proposed in~\cite{murai2024mast3r,maggio2025vggt}. 
Specifically, for methods that assume calibrated cameras, we estimate the intrinsic parameters from the first frame of each sequence using GeoCalib~\cite{veicht2024geocalib}, a learning-based single-image calibration approach.

 The results reported in Tables~\cref{tab:replica,tab:replica_results,tab:7scenesate,tab:tumrgbd} are all obtained under this evaluation protocol.
All baselines are run with their default or recommended settings as specified in the original publications.

\subsubsection{Metrics}~ We evaluate the proposed method on accuracy and efficiency. Accuracy is measured by RMSE-ATE (root-mean-square error of absolute trajectory error) [m], 
where the camera poses are estimated from corrected and optimized keypoints. Efficiency is evaluated in terms of average frame rate (FPS) across sequences and GPU memory (VRAM, video random-access memory) usage during sequence execution. We highlight the top-2 rankings, with bold indicating the best result and underline indicating the second best in each column.
\begin{table*}[ht]
\centering
\caption{Average runtime of the key modules in our system (ms).}
\label{tab:runtime_all_ms}
\scriptsize
\setlength{\tabcolsep}{2pt}  
\renewcommand{\arraystretch}{0.95}
\begin{tabular}{c|c|c|cccc|ccc|ccc|ccccc}
\hline
\multirow{2}{*}{Dataset} & \multirow{2}{*}{Sequence} &
\multicolumn{1}{c|}{Prepare} &
\multicolumn{4}{c|}{Per-frame tracking} &
\multicolumn{3}{c|}{Per-keyframe} &
\multicolumn{3}{c|}{Dense mapping} &
\multicolumn{5}{c}{Summary} \\
\cline{3-18}
& &
\makecell{Load\\Frame} &
\makecell{Feature\\extraction} &
\makecell{Feature\\matching} &
\makecell{Pose\\estimation} &
\makecell{Total} &
\makecell{Triangu-\\lation} &
\makecell{Local\\Loop} &
\makecell{Total} &
\makecell{Feed-forward\\Reconstruction} &
\makecell{Point Cloud\\registration} &
\makecell{Total} &
\makecell{Total\\frame} &
\makecell{Keyframe\\number} &
\makecell{Submap\\number} &
\makecell{Total\\time (s)} &
\makecell{FPS} \\
\hline
7-Scenes & chess    & 10.98 & 4.28 & 1.06 & 4.02 &  9.36 & 1.33 &  9.91 & 11.24 & 250.65 & 55.60 & 306.25 & 1000 & 33 &12& 27.6 & 36.13 \\
Replica  & room0    & 12.40 & 3.60 & 0.70 & 3.20 &  7.50 & 1.13 & 14.50 & 15.63 & 236.40 & 67.88 & 304.28 & 2000 & 37 & 11&44.23 & 45.21 \\
TUM      & fr1/room &  9.98 & 6.81 & 1.86 & 4.22 & 12.89 & 1.99 & 12.68 & 14.67 & 293.33 & 61.76 & 355.09 & 1362 & 163 &52& 43.64 & 31.20 \\
\hline
\end{tabular}
        \vspace{-4mm}

\end{table*}

\subsubsection{Quantitative Analysis}~\label{subsec:pose_estimation_eval}We present our camera pose estimation performance on the TUM-RGBD, 7-Scenes, and Replica datasets in Tables~\ref{tab:tumrgbd}, \ref{tab:7scenesate}, and \ref{tab:replica}. 

On the synthetic Replica dataset, our method achieves the lowest RMSE on the \textit{room0} and \textit{room1} sequences, with a slightly higher average RMSE across all sequences compared to MASt3R-SLAM and VGGT-SLAM. 
On the TUM-RGBD dataset, our method performs competitively in small-scale environments such as \textit{desk}, \textit{xyz}, and \textit{rpy}, while demonstrating superior accuracy in larger-scale sequences—including \textit{room}, \textit{floor}, and \textit{360}—where it consistently ranks among the top three methods. 
This improvement is attributed to our global loop closure mechanism, which effectively reduces trajectory drift and improves long-term pose consistency. 
The robust performance across scenes of varying scale and complexity underscores the strong generalization ability and robustness of our approach. 

Our method also achieves competitive results on the 7-Scenes dataset, with accuracy comparable to state-of-the-art methods. 
Most importantly, across all three benchmarks, our method achieves consistently low pose estimation errors while maintaining high computational efficiency—operating at high frame rates and consuming significantly less GPU memory than most existing approaches. 
This favorable balance between accuracy and efficiency demonstrates the practicality and effectiveness of our method for real-time monocular SLAM.

\begin{table}[ht]
\centering
\scriptsize
\renewcommand{\arraystretch}{0.9}
\setlength{\tabcolsep}{4pt}
\caption{Runtime Analysis on Multiple Devices}
\label{tab:runtime-analysis}
\begin{tabular}{lllccc}
\toprule
\textbf{Dataset} & \textbf{Sequence} & \textbf{Method} & \textbf{PC} & \textbf{Laptop} & \textbf{Jetson} \\
\midrule
\multirow{4}{*}{TUM-RGBD} & \multirow{4}{*}{fr1/room} 
& DROID-SLAM  & 23 & Failed  & 1   \\
& & MASt3R-SLAM & 13 & 2  & 0.1 \\
& & Ours(w.F3R)   & 27 & 15 & 7   \\
& & VGGT-SLAM   & 34 &Failed  &Failed   \\
\midrule
\multirow{4}{*}{7-Scenes} & \multirow{4}{*}{seq-01/chess} 
& DROID-SLAM  & 24  & Failed & 1   \\
& & MASt3R-SLAM & 17  & 2  & 0.1   \\
& & Ours(w.F3R)   & 32  & 18  & 9   \\
& & VGGT-SLAM   & 34 & Failed  &Failed   \\
\bottomrule
\end{tabular}
        \vspace{-4mm}

\end{table}

\subsection{Dense Reconstruction Evaluation}
\label{subsec:dense_reconstruction_eval}
Following the protocol of~\cite{slam3r,wang2025vggt,wang2024spann3r}, we construct a ground-truth point cloud for each test sequence by back-projecting RGB pixels into 3D space using corresponding ground-truth depth maps and camera poses. 
The resulting reconstructed point clouds are then aligned to the ground-truth models using the Umeyama algorithm~\cite{umeyama1991least} for closed-form similarity transformation estimation, followed by fine registration via Iterative Closest Point (ICP)~\cite{besl1992method}.

\subsubsection{Benchmark on 7-Scenes}~ 
Following~\cite{murai2024mast3r,maggio2025vggt}, we evaluate reconstruction in terms of the RMSE of accuracy, completeness, and symmetric Chamfer Distance.
As shown in Table~\ref{tab:7scenes_recon}, our method achieves highly competitive reconstruction performance, particularly excelling in accuracy. 
Compared to VGGT-SLAM, which is also based on the VGGT, our approach reduces the accuracy error by half (0.025 vs. 0.052).

\subsubsection{Benchmark on Replica}~ Following the evaluation protocols in~\cite{wang2024spann3r,slam3r,zhang2024hi,sandstrom2025splat}, we evaluate 3D reconstruction quality using mean accuracy and completeness, measured in centimeters (cm). In addition to the previously introduced state-of-the-art SLAM baselines, we further compare with recent 3D reconstruction methods: SLAM3R~\cite{slam3r}, DUSt3R~\cite{wang2024dust3r}, MASt3R~\cite{duisterhof2024mast3r}, and Spann3R~\cite{wang2024spann3r}.
As shown in Table~\ref{tab:replica_results}, our method achieves state-of-the-art performance, surpassing all existing approaches in both accuracy and completeness. 
Although VGGT has been pretrained on the Replica dataset, VGGT-SLAM, which builds directly on VGGT, still performs poorly on Replica (Acc: 6.32 vs. 2.82; Comp: 4.01 vs. 2.07). In contrast, our framework achieves significantly better accuracy, highlighting its effectiveness beyond merely leveraging VGGT. Moreover, when applied with Fast3R as the mapping backbone, our system also delivers strong results, demonstrating the robustness and versatility of the proposed framework.
\subsection{Time Analysis}
As shown in Table~\ref{tab:runtime_all_ms}, our per-frame tracking runs at an exceptionally high speed thanks to the use of a lightweight neural network. 
This design choice also explains why our system is substantially more efficient than methods such as MASt3R-SLAM, Spann3R, and SLAM3R. Although the mapping module remains relatively time-consuming, it is executed in a separate thread and therefore does not affect tracking performance.  In addition, the significantly higher runtime speed observed on the Replica dataset is mainly attributed to the smoother camera motion, which results in fewer detected keyframes.
\begin{figure*}[t]
    \centering
    \includegraphics[width=\textwidth]{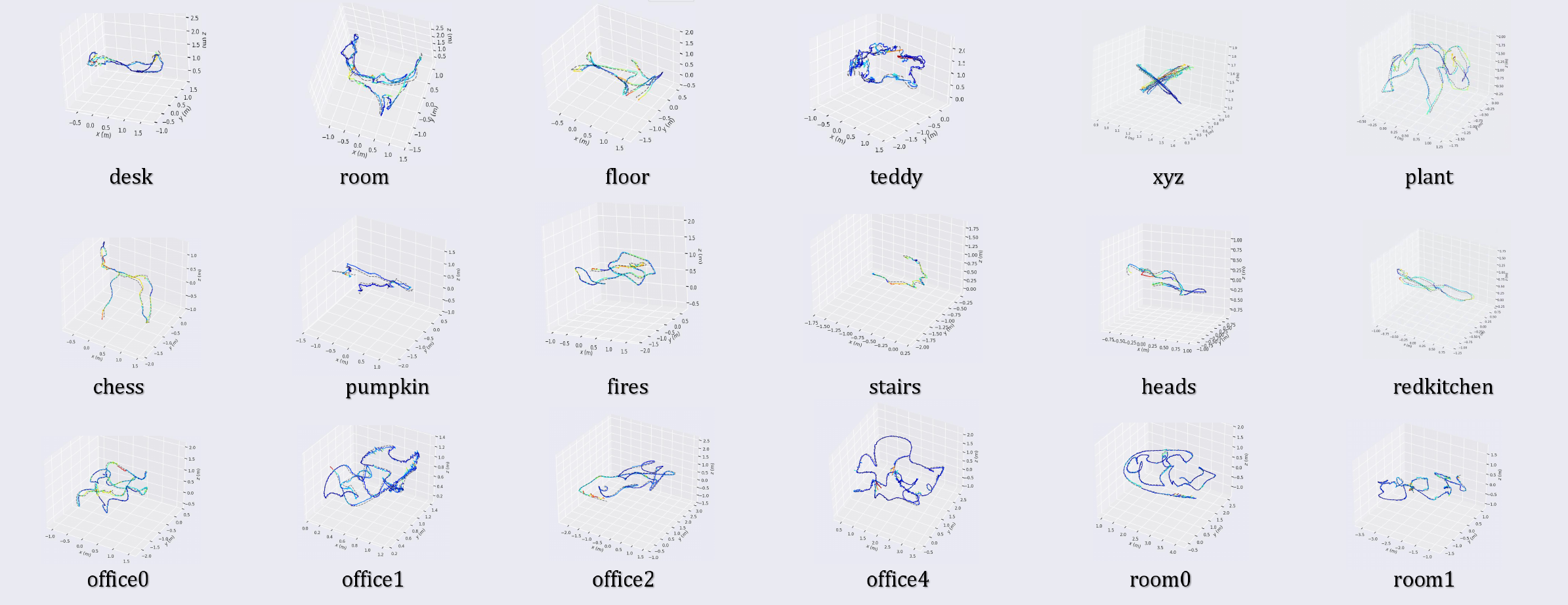}
    \caption{Qualitative trajectory visualization on TUM-RGBD, 7-Scenes, and Replica datasets. 
    Blue lines denote the ground-truth trajectories and green lines indicate the estimated trajectories from our method.}
    \label{fig:trajectory}
\end{figure*}
\subsection{Qualitative Results}
\label{subsec:qualitative_results}
As shown in Figure~\ref{fig:trajectory}, our method produces trajectory estimates that closely align with the ground truth across multiple datasets, including TUM-RGBD, 7-Scenes, and Replica. These results highlight the robustness of our framework in handling diverse environments and sensor conditions. Compared with existing approaches, our trajectories exhibit reduced drift and better alignment with the reference paths, thereby demonstrating both accurate pose estimation and strong generalization capability. This generalization ability is further confirmed in Figure~\ref{fig:generalization}, where our method consistently adapts to different datasets without requiring retraining. Moreover, Figure~\ref{fig:qual_replica_tum} and~\ref{fig:qual_replica_7scenes} illustrates that our approach achieves superior multi-view consistency in reconstruction compared to competing methods, leading to more coherent and reliable 3D scene representations.
\subsection{Run on Multiple Devices}
We include Ours (w. Fast3R) in the comparison, since it can run under the 8 GB VRAM constraint of the laptop. We then compare our method with state-of-the-art monocular dense SLAM approaches across diverse hardware platforms on real-world datasets. As shown in Table~\ref{tab:runtime-analysis}, evaluations are conducted on a laptop, an NVIDIA Jetson Orin NX, and a desktop PC. The laptop is equipped with an NVIDIA RTX 4060 GPU (8 GB VRAM) and an Intel i5-13500H CPU.

\begin{table}[ht]
    \centering
    \scriptsize
    \setlength{\tabcolsep}{4pt}
    \caption{Ablation Study on Loop Closure Configurations}
    \label{tab:ablation-loop}
    \resizebox{\linewidth}{!}{
    \begin{tabular}{cc|c|ccc|ccc}
        \hline
        \multirow{2}{*}{\textbf{Local}} & 
        \multirow{2}{*}{\textbf{Global}} & 
        \textbf{TUM} & 
        \multicolumn{3}{c|}{\textbf{7-Scenes}} & 
        \multicolumn{3}{c}{\textbf{Replica}} \\
        \cline{3-9}
        & & \textbf{ATE$\downarrow$} & \textbf{ATE$\downarrow$} & \textbf{Acc$\downarrow$} & \textbf{Comp$\downarrow$} & \textbf{ATE$\downarrow$} & \textbf{Acc$\downarrow$} & \textbf{Comp$\downarrow$} \\
        \hline
        $\times$ & $\times$ & 0.225 & 0.185 & 0.048 & 0.069 & 0.084 & 0.042 & 0.033 \\
        $\times$ & $\checkmark$ & \underline{0.122} & 0.153 & 0.033 & \underline{0.056} & 0.066 & 0.031 & \underline{0.023} \\
        $\checkmark$ & $\times$ & 0.206 & \underline{0.096} & \underline{0.028} & 0.058 & \underline{0.043} & \textbf{0.025} & 0.025 \\
        $\checkmark$ & $\checkmark$ & \textbf{0.070} & \textbf{0.075} & \textbf{0.025} & \textbf{0.054} & \textbf{0.041} & \underline{0.028} & \textbf{0.021} \\
        \hline
    \end{tabular}
    }

\end{table}

\label{subsec:ablation_studies}
\begin{figure}[ht]
    \centering
    \includegraphics[width=0.9\linewidth]{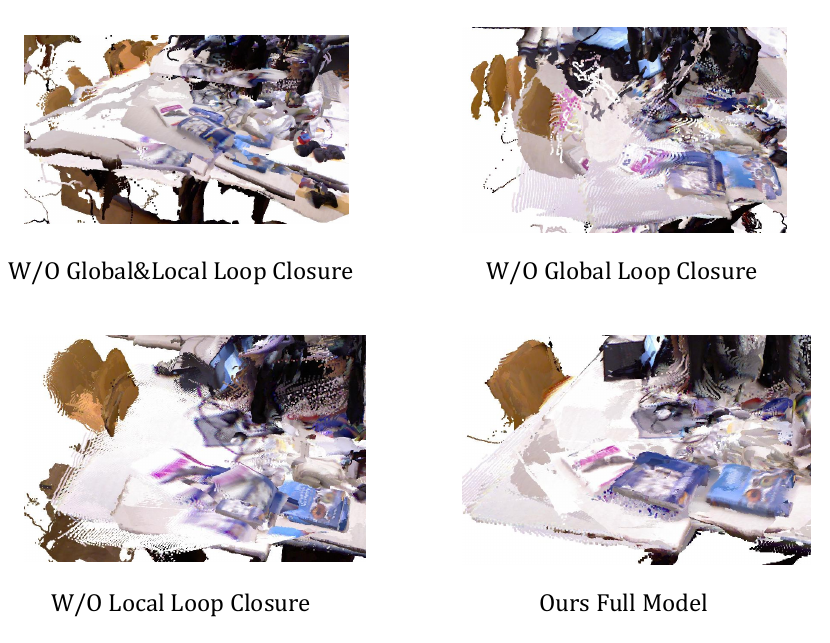}
    \vspace{-2mm}
    \caption{Ablation study on the TUM-RGBD desk2 sequence. }
    \label{fig:ablation_tum_desk2}
            \vspace{-4mm}

\end{figure}
\subsection{Ablation Studies}

We conduct an ablation study to evaluate the impact of local and global loop closure modules on trajectory accuracy, measured by Absolute Trajectory Error (ATE) RMSE across three benchmark datasets: TUM, 7-Scenes, and Replica. As shown in Table~\ref{tab:ablation-loop}, the baseline without any loop closure achieves the highest ATE values, indicating significant drift over time.

Introducing global loop closure alone reduces ATE by up to 46\% on TUM and 25\% on Replica, demonstrating its effectiveness in correcting long-term drift through global consistency optimization. However, it performs slightly worse than the baseline on 7-Scenes, suggesting that global constraints may be less effective in small-scale environments with limited loop opportunities.

Adding local loop closure further improves performance, reducing ATE by 10\% on TUM and 35\% on Replica compared to the global-only case. This highlights the importance of short-term geometric consistency in refining pose estimates.

The best results are achieved when both local and global loop closures are enabled, achieving ATE RMSEs of 0.062~m, 0.072~m, and 0.051~m on TUM, 7-Scenes, and Replica respectively—representing a 70\% reduction from the baseline. This confirms that the synergistic combination of local and global loop closure effectively suppresses both short-term noise and long-term drift, leading to robust and accurate localization.
As shown in Figure~\ref{fig:ablation_tum_desk2}, the visualization of reconstruction on the TUM-RGBD \textit{desk2} sequence demonstrates that by incorporating both local and global loop closure, our method is able to achieve multi-view consistent reconstruction results.

\section{Conclusion}
In this work, we proposed EC3R-SLAM, an efficient and consistent monocular dense SLAM system that integrates feed-forward 3D reconstruction with both local and global loop closure. 
Comprehensive evaluations on TUM-RGBD, 7-Scenes, and Replica datasets demonstrate that our method achieves superior accuracy compared to state-of-the-art uncalibrated dense SLAM approaches, while maintaining lower memory consumption and competitive real-time performance.



\bibliographystyle{IEEEtran}
\bibliography{bibliography/papers}

\end{document}